\documentclass[runningheads]{llncs}
\usepackage[T1]{fontenc}
\usepackage{hyperref}
\usepackage{color}

\usepackage[dvipsnames]{xcolor}

\usepackage{bm}
\usepackage{booktabs}
\usepackage{multirow}
\usepackage{siunitx}
\usepackage{subcaption}
\def\tablevspace{\vspace{9pt}}
\usepackage{microtype}
\usepackage{graphicx}
\usepackage{booktabs} 
\usepackage{xcolor}
\usepackage{hyperref} 
\usepackage{amsmath,amssymb}
\usepackage{bm}
\usepackage{booktabs}
\usepackage{comment}
\usepackage{multirow}
\usepackage{framed}

\usepackage{url}
\makeatletter
\newcommand{\figcaption}[1]{\def\@captype{figure}\caption{#1}}
\newcommand{\tblcaption}[1]{\def\@captype{table}\caption{#1}}
\makeatother

\newcount\K
\def\bla#1{
\K=0 \loop\ifnum\K<#1
{\textcolor[gray]{0.9}{{\it bla bla bla bla bla bla bla bla bla bla bla bla bla bla bla}}}
\advance\K by1\repeat
}

\newcommand{\todo}[1]{
\ifx#10
\textcolor{red}{$0.00_{\pm 0.00}$}
\else
\textcolor{red}{#1}
\fi
}

\begin{document}
\title{On the Relationship Between Double Descent of CNNs and Shape/Texture Bias Under Learning Process}
\titlerunning{CNN Learning: Double Descent vs. Shape/Texture Bias}
%
\author{
Shun Iwase\inst{1}\orcidID{0009-0007-3610-0109} \and
Shuya Takahashi\orcidID{0009-0006-3886-0825} \and
Nakamasa Inoue\inst{2}\orcidID{0000-0002-9761-4142} \and
Rio Yokota\inst{2}\orcidID{0000-0001-7573-7873} \and
Ryo Nakamura\inst{3}\orcidID{0009-0009-0858-1229} \and
Hirokatsu Kataoka\inst{4}\orcidID{0000-0001-8844-165X} \and
Eisaku Maeda\inst{1}\orcidID{0009-0005-0614-4923}
}
\authorrunning{S. Iwase et al.}
%
\institute{
Tokyo Denki University, Tokyo, Japan
\email{23amj03@ms.dendai.ac.jp}
\email{maeda.e@mail.dendai.ac.jp}
\and
Tokyo Institute of Technology, Tokyo, Japan
\and 
Tenchijin Inc., Tokyo, Japan
\and
National Institute of Advanced Industrial Science and Technology, Ibaraki, Japan
}
\maketitle              
\begin{abstract}
The double descent phenomenon, which deviates from the traditional bias-variance trade-off theory, attracts considerable research attention; however, the mechanism of its occurrence is not fully understood. On the other hand, in the study of convolutional neural networks (CNNs) for image recognition, methods are proposed to quantify the bias on shape features versus texture features in images, determining which features the CNN focuses on more. In this work, we hypothesize that there is a relationship between the shape/texture bias in the learning process of CNNs and epoch-wise double descent, and we conduct verification. As a result, we discover double descent/ascent of shape/texture bias synchronized with double descent of test error under conditions where epoch-wise double descent is observed. Quantitative evaluations confirm this correlation between the test errors and the bias values from the initial decrease to the full increase in test error. Interestingly, double descent/ascent of shape/texture bias is observed in some cases even in conditions without label noise, where double descent is thought not to occur. These experimental results are considered to contribute to the understanding of the mechanisms behind the double descent phenomenon and the learning process of CNNs in image recognition.

\end{abstract}

\section{Introduction} Deep learning has become an important research area with applications in many fields such as computer vision. To build high-performance models with deep learning, it is essential to prepare appropriate training datasets and determine the number of model parameters according to these datasets. Two crucial phenomena to understand in this context are underfitting\cite{underfitting} and overfitting\cite{overfitting}. Underfitting occurs even with sufficient training data if there are too few model parameters and prevents the model from learning features from the training data, leading to suboptimal performance on test data. Conversely, overfitting occurs when there is either too little training data or an excess of model parameters, causing the model to fit too closely to the training data and hindering generalization to new test data. Researchers widely recognize these phenomena as the bias-variance trade-off\cite{Rajnarayan2010}.

However, recent studies report an interesting phenomenon: when the number of model parameters becomes very large, performance can improve again after overfitting\cite{Belkin_2019,nakkiran2021deep,Advani,dar2020subspace}. Belkin et al. name this phenomenon "double descent" and demonstrate it in models such as two-layer neural networks and random forests.\cite{Belkin_2019} Later, Nakkiran et al. confirm this phenomenon in more practical deep neural networks such as convolutional neural networks (CNNs) for image classification.\cite{nakkiran2021deep} Moreover, they show that in addition to increasing the number of model parameters, increasing the training epochs can also induce double descent. Many studies on double descent focus on elucidating its theoretical foundations. Particularly, research on double descent due to an increase in the number of training epochs considers the features of the data and how models learn these features~\cite{Pezeshki,Stephenson,Heckel}. However, such investigations mainly use artificial data or non-deep learning models, and studies centered on the characteristics of real data in deep learning, such as shape and texture specific to images, have rarely been conducted.

\begin{figure}[t]
\centering
\includegraphics[width=0.75\columnwidth]{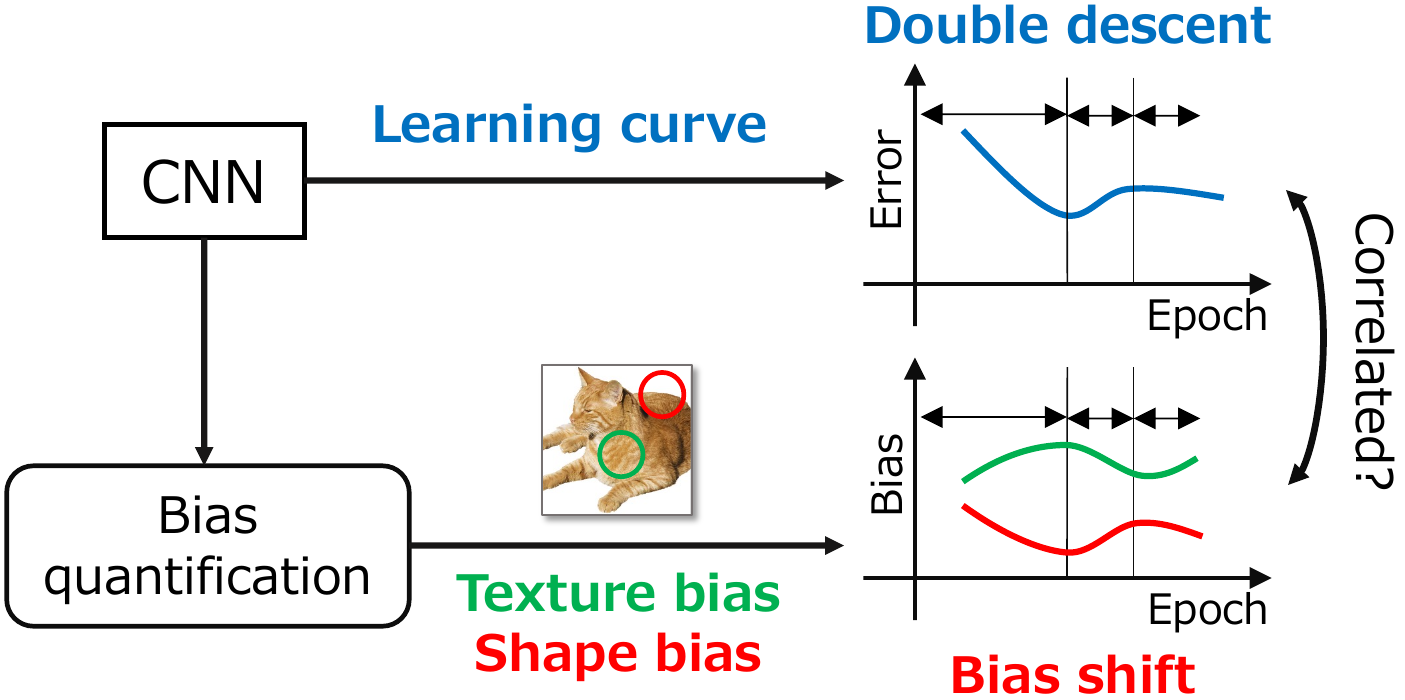}
\vspace{-12pt}
\caption{
Flow of analysis process presented in this paper. We train CNNs for image recognition under double descent conditions. We monitor the temporal evolution of the shape/texture bias and test error assessing the capacity of the model to interpret shapes and textures while exploring their correlation.
}
\label{fig:fig2}
\vspace{-8pt}
\end{figure}

On the other hand, in image classification, analyses focusing on features such as shape and texture reveal that CNN models trained on ImageNet tend to be biased towards texture features~\cite{Geirhos}. It has also been found that this texture bias can be reduced using simple data augmentation and prolonged training, leading to a stronger bias towards shape features~\cite{Hermann}. Given these characteristics of CNNs, important questions arise about the relationship between the double descent phenomenon and learning of shape and texture features with CNNs. However, a comprehensive exploration of the relationship between the learning phases for shape and texture and the various phases of double descent has not yet been conducted.

In this study, we delve into the relationship between image-specific features (shape and texture) and the double descent of CNNs. The flow of analysis is shown in Fig.~\ref{fig:fig2}. First, we define the period until the initial increase in test error as Phase 1, the period until it starts to decrease as Phase 2, and the subsequent period as Phase 3. Second, we assess the relationship between test error and the bias toward shape and texture during these phases. Specifically, we quantify the shape/texture bias of the CNN model by utilizing the method proposed by Islam et al. \cite{Islam} during training, and compare the trajectories of the bias with the progression of double descent. Furthermore, we calculate the correlation coefficients between the test errors and the bias values in each of the three phases for a more quantitative evaluation. We also conduct ablation studies and analyses under various conditions. The contributions of the present paper are as follows: \begin{figure}[t]
\centering
\includegraphics[width=0.8\linewidth]{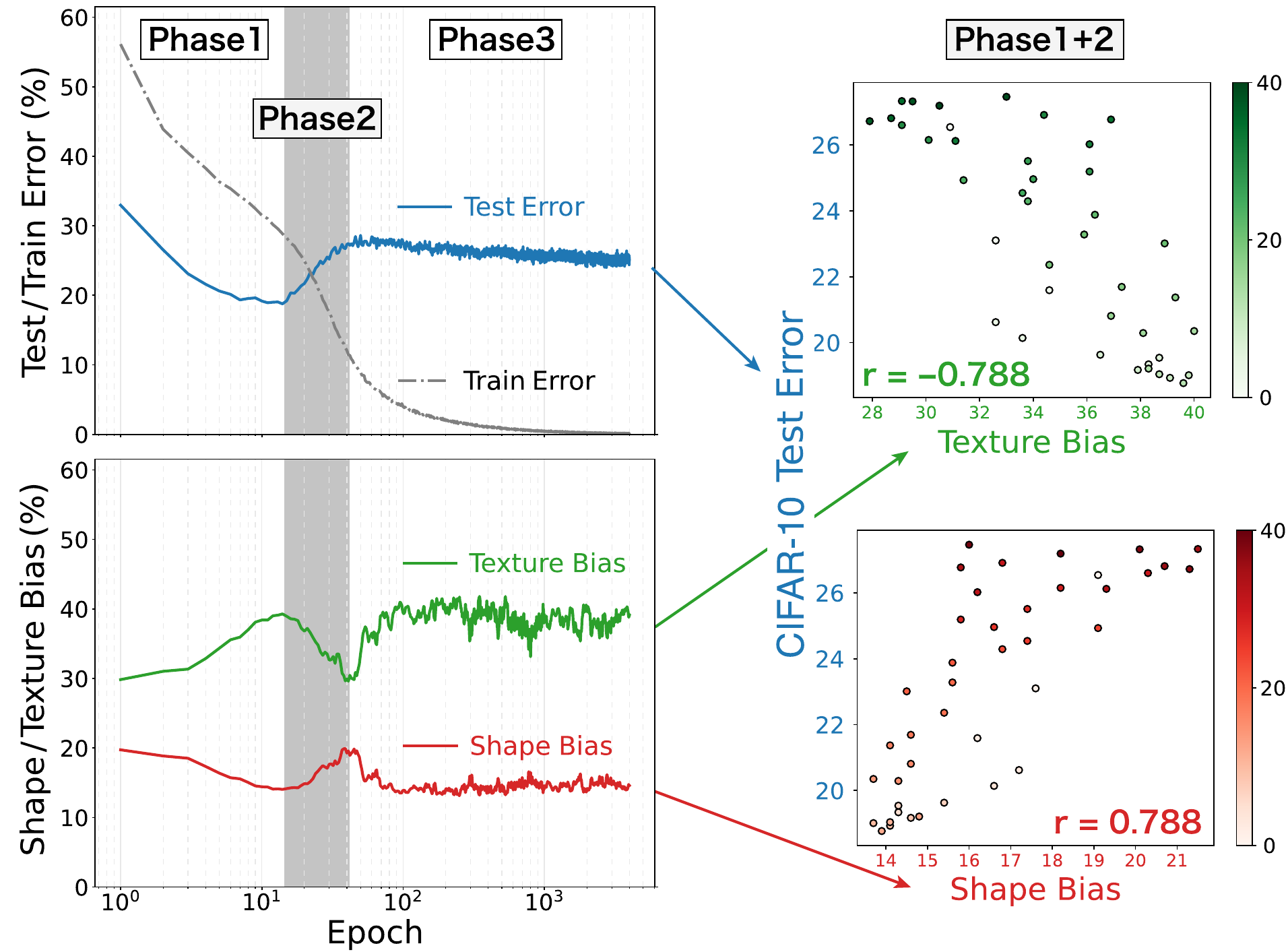}
\caption{
Schematic overview of this study. Top left: The learning curve of the CIFAR-10 image recognition task under the setting of \cite{nakkiran2021deep} et al,
where epoch-wise double descent was observed. Test errors were divided into three phases based on their temporal differentiation. Bottom left: This records the model's shape/texture bias during the aforementioned learning process. It shows the synchronous changes between test errors and shape/texture bias. Right: A scatter plot of test error and shape/texture bias. Especially in Phase 1 and Phase 2, there is a positive correlation between test error and shape bias, and a negative correlation between test error and texture bias. In all bias visualization settings, including this one, we use a 5-term moving average to smooth the data for trend analysis.
}
\label{fig:overview}
\end{figure}
 { \setlength{\leftmargini}{12pt} \begin{itemize} \item We conduct the first analysis of features of natural images, such as shape and texture, in the context of the double descent phenomenon in deep learning. We calculate the model's bias toward shape and texture and compare the evolution of this bias with changes in test errors throughout the learning process.

\item As a result, we find that shape/texture bias and test error are often correlated in Phases 1 and 2. In Fig.  \ref{fig:overview}, a strong correlation is present in Phases 1 and 2, while Phase 3 tends to show no correlation, under Nakkiran's setting. Interestingly, the inflection points in the temporal progression of the test error and the shape/texture bias almost coincide. To the best of our knowledge, we are the first to report the double descent/ascent phenomena of shape/texture bias and its synchronization with the double descent phenomena of test error.

\item To better understand the phenomena, we perform ablation studies and analyses beyond Nakkiran's setting, including experiments with various CNN architectures and different noise levels. One interesting finding is that we observe double descent/ascent of shape/texture bias even without adding noise to the labels, a condition where double descent is thought not to occur (Fig.  \ref{fig:comp_ln}).

\end{itemize} }

\section{Related Work}
\subsection{Double descent}


A recently discovered phenomenon called double descent~\cite{Belkin_2019} shows that as model complexity increases further, performance improves again. In other words, after the initial U-shaped curve (transition from underfitting to overfitting), a new phase of performance improvement appears with increased complexity. Over-parameterized deep neural networks, theoretically prone to overfitting, sometimes demonstrate superior generalization performance~\cite{He:ResNet,ViT,NEURIPS2020_1457c0d6}.
Belkin et al. \cite{Belkin_2019} first confirmed double descent in decision trees and two-layer neural networks. Later, Nakkiran et al. \cite{nakkiran2021deep} showed that it also occurs in deep neural networks and with more learning epochs. Reports also indicate double descent happens with increased sparsity due to parameter pruning \cite{He}. Double descent observed with more parameters, learning epochs, and increased sparsity is called model-wise double descent, epoch-wise double descent, and sparse double descent, respectively \cite{nakkiran2021deep,He}.
%
The discovery of these phenomena challenges traditional interpretations related to the design and parameter selection of models. It significantly impacts both the theory and practice of machine learning. Understanding and utilizing these phenomena could contribute to the development of more efficient and versatile machine-learning models.

\textbf{Model-wise double descent.}
Yang et al. \cite{Yang} revisited the classic theory of the bias-variance trade-off through extensive experiments. They found that while bias monotonically decreases as classification theory predicts, variance shows unimodal behavior. This combination of bias and variance suggests three typical risk curve patterns, aligning with many already reported experimental results.

\textbf{Epoch-wise double descent.}
Several hypotheses about double descent in the learning process emerge from statistical simulation results. These hypotheses focus on the characteristics of the data. For example, Stephenson et al. \cite{Stephenson} assume that double descent occurs due to slow yet beneficial features and show that removing the principal components of data in an ideal linear model can eliminate the double descent behavior. On the other hand, Pezeshki et al. \cite{Pezeshki} find through experiments that features learned at different scales cause double descent. Moreover, Heckel et al. \cite{Heckel} state that overlapping trade-offs between multiple biases and variances, due to different parts of the model learning at different epochs, trigger double descent. They demonstrate that varying learning rates across layers can mitigate double descent.

\textbf{Sparse double descent.}
As the model's sparsity increases, meaning many parameters become zero or very small, we first observe performance improvement. However, performance declines after a certain point. Further increasing sparsity, performance improves again \cite{He,SDD_VIT}. This suggests that moderate sparsity, achievable through methods like network pruning, can suppress model overfitting and enhance generalization performance.

\subsection{CNN for image understanding}
Geirhos et al. \cite{Geirhos} showed that CNNs trained on ImageNet especially emphasize image textures for classification. The input images with conflicting shape and texture information into CNNs and checked whether the output matched shape-based or texture-based labels. Based on these results, they analyzed whether CNNs prioritize shape or texture in recognition. Meanwhile, Islam et al. \cite{Islam} proposed a method to quantitatively determine the emphasis on shape and texture in models based on neurons' latent representations. This method allowed them to analyze which features CNNs emphasize or ignore. Furthermore, Ge et al. \cite{Ge} attempted to model the human visual system and developed the Human Vision System (HVS). HVS can quantitatively evaluate which features (shape, texture, color, etc.) play the most crucial role during image classification. Our research builds on prior studies about CNNs in image understanding and double descent. We attempted to reveal the relationship between the acquisition of knowledge about texture and shape information during CNN learning and the phenomenon of double descent.

\begin{figure}[t]
\centering
\includegraphics[width=0.8\textwidth]{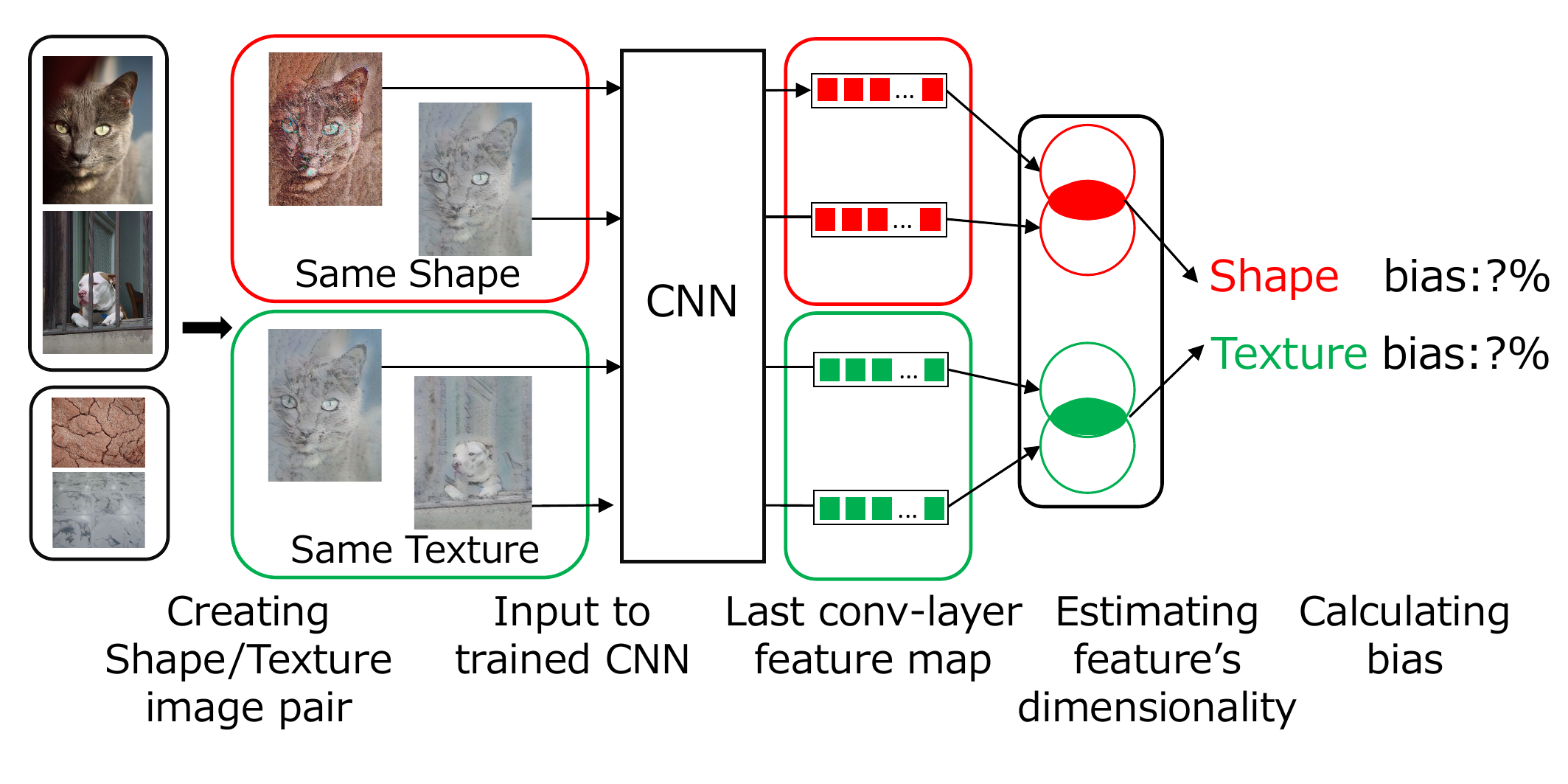}
\caption{Overview of Islam's method. This figure shows an overview of the process of calculating the shape and texture bias using the method of Islam et al. \cite{Islam}}
\label{fig:fig3}
\end{figure}

\section{Correlation analysis framework of double descent and shape/texture bias}

This section explains how to investigate the relationship between epoch-wise double descent in deep learning and the shape and texture features of natural images. Figure \ref{fig:fig2} outlines this method. First, we train a CNN under conditions that cause double descent and observe the progression of the learning curve. Additionally, we quantify the bias towards shape and texture features at each epoch using the method of Islam et al. \cite{Islam} and similarly observe its progression during training. By doing so, we compare the progression of double descent and bias. Furthermore, for quantitative evaluation, we divide double descent into three phases and assess the correlation coefficient between test error rate and shape/texture bias in each phase. The following sections explain the observation method for epoch-wise double descent, the division of its phases, and the calculation method for shape and texture bias.

\subsection{How to observe double descent}
To observe epoch-wise double descent, we use the conditions originally used by Nakkiran et al. \cite{nakkiran2021deep}. They have observed double descent under various conditions. Among these conditions, we adopt the setup involving ResNet18~\cite{He:ResNet} and CIFAR-10~\cite{Krizhevsky2009_cifar} as the baseline condition, for our study.  In this condition, they added noise to the labels of the training data. This addition makes the double descent more pronounced.

\subsection{Phases of learning curve with double descent}
\label{sec:Phase division of double descent} 
In this study, in order to analyze the relationship between double descent and the model's shape/texture bias, we divide the learning process into the following three phases based on test error.
\textbf{Phase 1}: From the beginning of training until the test error reaches its minimum.
\textbf{Phase 2}: From the end of Phase 1 until the test error decreases again.
\textbf{Phase 3}: From the end of Phase 2 thereafter.

To determine these phases, we utilize a gradient-based method. Specifically, we monitor the test error across epochs and compute the difference $\Delta e$ between consecutive epochs as $\Delta e=\left|e_i-e_{i+5}\right|$. At any given epoch $i$, if the difference $\Delta e$ is less than or equal to a specified threshold $\theta$, the test error has either stabilized or has improved slightly. We define the interval up to the smallest epoch number at this time as Phase~1. For Phase~2, encompasses the interval from the epoch number just after Phase~1 to the smallest epoch where the difference is less than or equal to $\theta$. Phase 3 refers to any interval following Phase 2. In our experiments, the threshold $\theta$ is set at 0.1.
In the experimental setup used, the phases are divided by the process described above because empirically the second descent of a double descent does not have a lower test error rate than the first descent.

\subsection{Quantifying the shape/texture bias of the model}
\label{sec;Quantifying the shape/texture biases of the model}
We estimate the number of neurons encoding shape and texture features in the final convolutional layer of CNNs using the method proposed by Islam et al. \cite{Islam}, and define this ratio as the model's shape and texture bias\footnote{https://github.com/islamamirul/shape\_texture\_neuron}. The flow for calculating shape and texture bias is shown in Fig. \ref{fig:fig3}.
For quantifying shape and texture bias, we use the Stylized PASCAL VOC 2012 (SVOC) dataset~\cite{pascal-voc-2012} created by the AdaIn transfer algorithm\cite{AdaIn} from the PASCAL VOC 2012 dataset and the Describable Textures Dataset~\cite{cimpoi14describing}. We sample image pairs with common shape and texture features from the SVOC dataset. Then, we calculate the correlation coefficients $\rho^{shape}_i$ and $\rho^{texture}_i$ for each feature. For example, we input image pairs with common shapes into the model and obtain outputs $z_i^a$ and $z_i^b$ from neuron $z_i$. We calculate the correlation coefficient $\rho^{shape}_i$ from these outputs $z_i^a$ and $z_i^b$. We follow a similar procedure to calculate $\rho^{texture}$.
We determine the proportion of neurons encoding shape and texture features (shape and texture bias) respectively, by calculating the softmax of the sum of each $\rho^{shape}_i$ and $\rho^{texture}_i$ and the baseline value (number of neurons $|z|$). For more details on the SVOC construction method and the concept of this technique, see Islam et al. \cite{Islam} .

\section{Experiments}
\label{sec:Experiment}
In this section, the following three sets of experiments are conducted:
1) Under the setting of Nakkiran et al.~\cite{nakkiran2021deep}, where epoch-wise double descent was confirmed, we compare the progression of test error rates and shape/texture bias. We also quantitatively investigate correlations in each phase defined in Fig. \ref{sec:Phase division of double descent}.
2) We conduct ablation studies and analyses to deepen the understanding of the relationship between the double descent of test error and shape/texture bias.
3) We conduct layer-wise analyses by evaluating the shape/texture bias of each layer and visualizing the filters of the first layer.

\subsection{Nakkiran's setting}
\label{sec:Experimental setting}
\label{sec46}
\textbf{Detailed settings.}
The ResNet18 model with weights pre-trained on ImageNet~\cite{Deng} is trained on CIFAR-10~\cite{Krizhevsky2009_cifar} using label noise and data augmentation.
The label noise involves randomly changing the correct label of the training data to another label with a probability of $p = 0.2$.
For data augmentation, random cropping and flipping are utilized. For cropping, a 4-pixel margin is added to the top, bottom, left, and right sides of the input image, and then the image is cropped to a size of $32 \times 32$. Flipping is applied horizontally. The batch size is set to 128. The cross-entropy loss is used as the loss function. For optimization, Adam~\cite{Adam} is used with a learning rate of $10^{-4}$.
This experimental condition is consistent with that of Nakkiran et al.~\cite{nakkiran2021deep}, where an epoch-wise double descent phenomenon was observed.

\begin{figure*}[t]
\begin{minipage}[b]{0.65\linewidth}
\centering
\includegraphics[width=\linewidth]{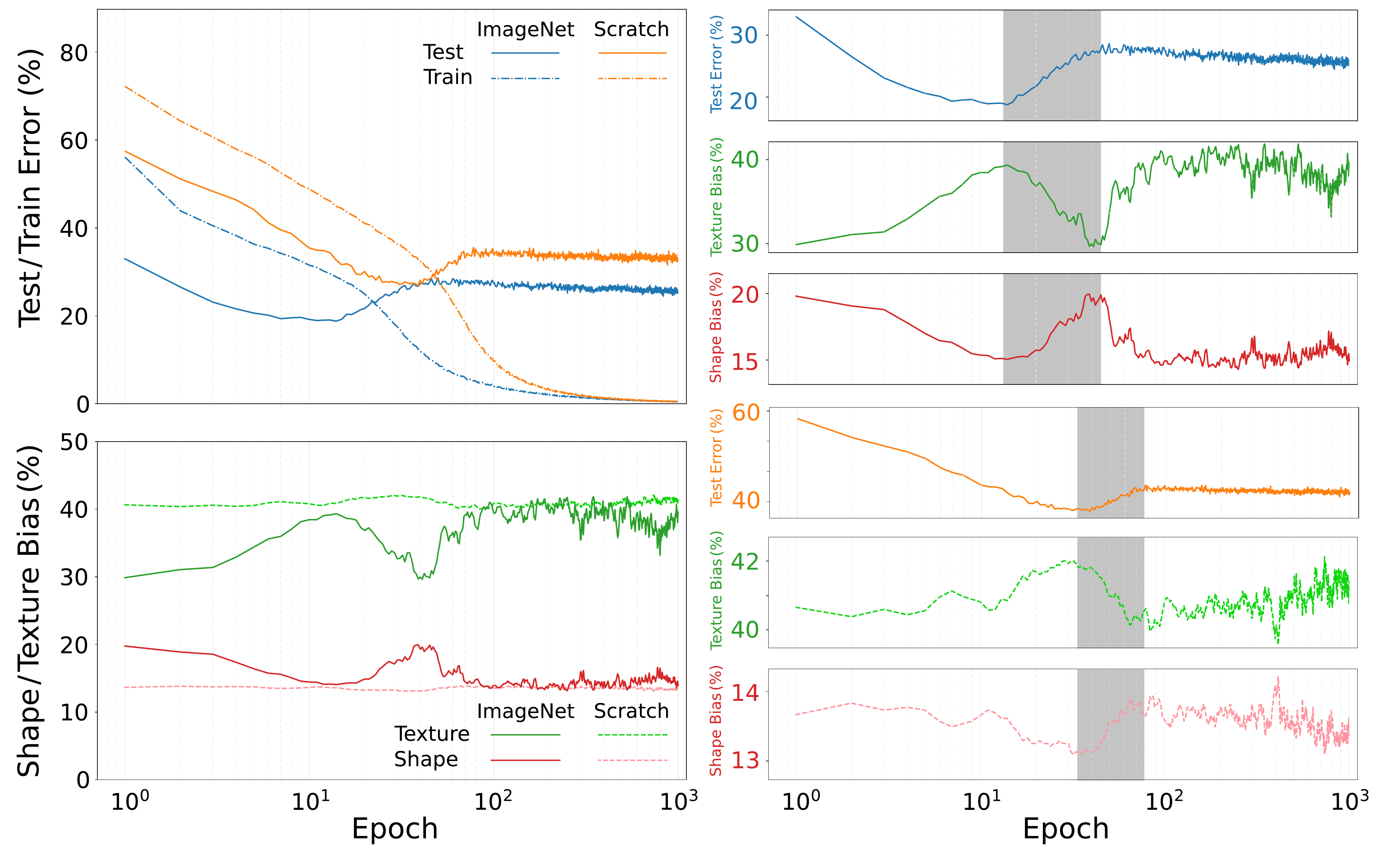}
\caption{Comparison of training with randomly initialized weights (Scratch) and with weights pre-trained on ImageNet (ImageNet). Left top: train and test errors (\%). Left bottom: shape/texture bias (\%).
Right: Enlarged view.}
\label{fig:comp_INSR}
\end{minipage}
\hfill
\begin{minipage}[b]{0.325\linewidth}
\centering
\includegraphics[width=\linewidth]{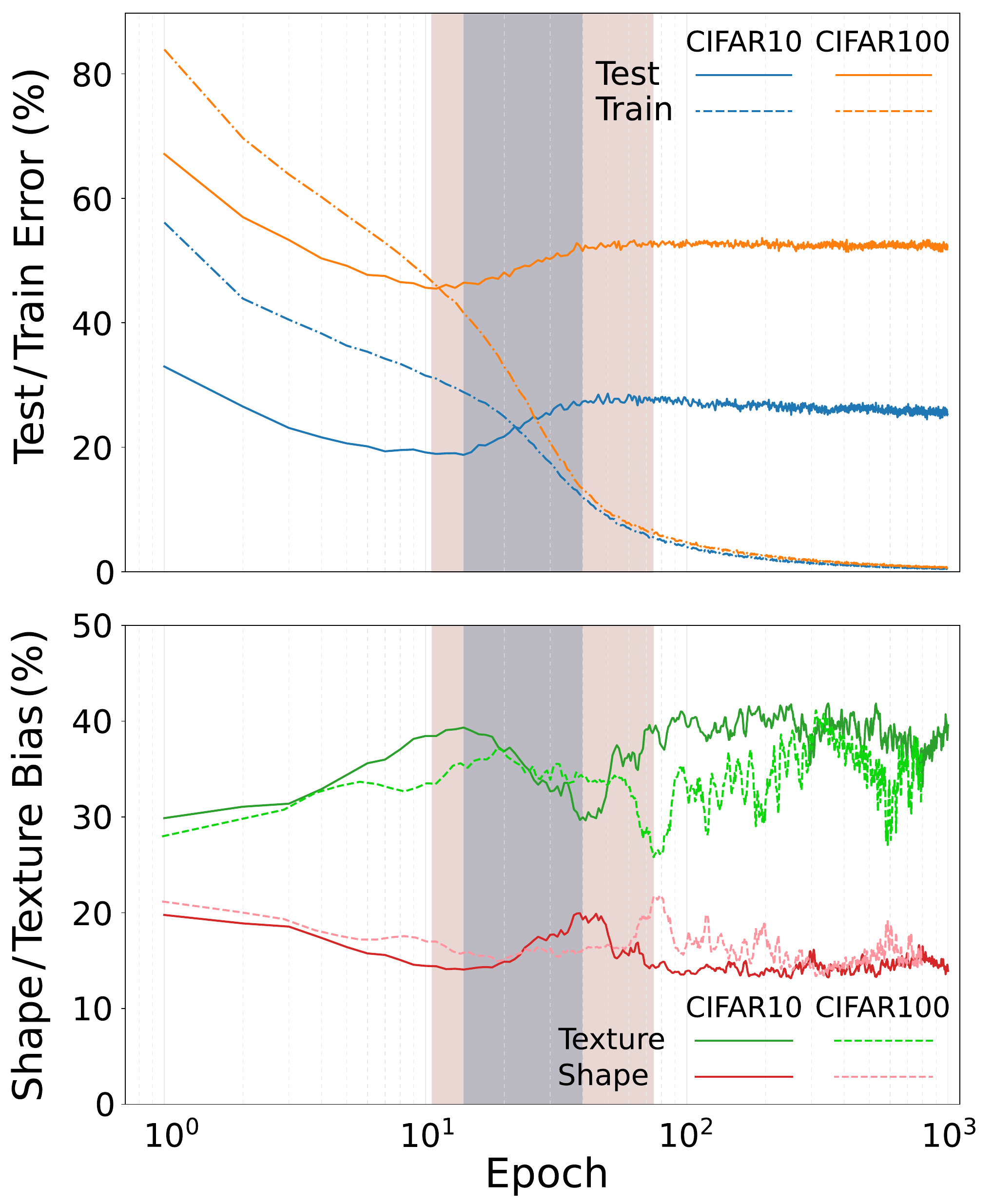}
\caption{Comparison of CIFAR-10 and CIFAR-100. Top: train and test errors (\%). Bottom: shape/texture bias (\%).
}
\label{fig:comp_IN_dataset}
\end{minipage}
\vspace{-10pt}
\end{figure*}
\label{sec:results}

\textbf{Experimental results.} The results comparing the progression of the test error and the shape/texture bias are shown in Fig. \ref{fig:overview}.
We observed double descent phenomena of test error and double descent/ascent phenomena of shape/texture bias.
Comparing the trends in the test error and the shape bias, a correlation is observed with a decrease in Phase 1, an increase in Phase 2, and a decrease again in Phase 3. There is also an inverse correlation between the test error and the texture bias.
The inflection points in the temporal progression of the test error and the shape/texture bias
almost coincide.

\textbf{Correlation analysis in each phase.}
We calculated the correlation coefficients between the test error and the shape/texture bias in Phases 1, 2, and 3 for a more detailed evaluation. The results are shown in Fig. \ref{tab:base_condition_result}. The correlation coefficient $r_{\text{shape}}$ for shape bias in Phases 1 and 2 were $0.898$ and $0.771$, respectively, indicating a positive correlation. Conversely, the correlation coefficient $r_{\text{texture}}$ for texture bias in Phases 1 and 2 were $-0.829$ and $-0.797$, indicating a negative correlation. In Phase 3, these correlation coefficients were $-0.026$ and $0.118$, respectively, showing no significant correlation. These results indicate that there are correlations in Phases 1 and 2, but not in Phase 3.

\begin{table}[t]
\centering
\setlength{\tabcolsep}{10pt}
\caption{
Correlation coefficients and synchronization scores. Phase: the three phases divided according to the method defined in \ref{sec:Phase division of double descent}.
Epoch range: The start and end epoch of each phase.
$r_{\text{shape}}$: correlation coefficients between shape bias and test error.
$r_{\text{texture}}$: correlation coefficients between texture bias and test error.
$s$: synchronization score.
}
\tablevspace
\scalebox{0.85}{
\begin{tabular}{c|c|cc|c}
\toprule
Phase & Epoch range & $r_{\text{shape}}$ & $r_{\text{texture}}$ & $s$\\
     \midrule
    Phase 1 & 2 - 12 &$0.898$ &$-0.829$ &$0.863$\\
    Phase 2 & 12 - 41&$0.771$& $-0.797$&$0.784$\\
    Phase 3 & 41 - 1,000&$-0.026$ & $0.118$&$0.072$\\
    \bottomrule
    \end{tabular}
}\label{tab:base_condition_result}
\end{table}

\textbf{Synchronization score.}
To simplify the evaluation, we introduce synchronization score $s$ defined by the average of the absolute values of the two correlation coefficients, \textit{i.e.,} $s = \frac{1}{2}(|r_{\text{shape}}| + |r_{\text{texture}}|)$. A higher score indicates a stronger synchronization between test error and shape/texture bias.
In Fig. \ref{tab:base_condition_result}, strong synchronization with scores greater than 0.7 is observed in Phases 1 and 2.

\subsection{Ablation studies and analyses}

To better understand the relationship between the double descent of test error and the double descent/ascent of shape/texture bias, we perform ablation studies and analyses beyond Nakkiran’s setting. Specifically, we conduct experiments with respect to parameter initialization, dataset selection, architecture and label noise.

\def\tabletemplate{
&  \multicolumn{3}{c|}{Correlation in Phase 1, 2} &  \multicolumn{3}{c}{Correlation in Phase 3}\\
& $r_{\mathrm{shape}}$ & $r_{\mathrm{texture}}$ & $s$ & $r_{\mathrm{shape}}$ & $r_{\mathrm{texture}}$ & $s$
}
\begin{table}[t]
\centering
\setlength{\tabcolsep}{4.3pt}
\caption{Correlation coefficients and synchronization scores for CIFAR-10 and CIFAR-100 datasets.}
\tablevspace
\scalebox{0.85}{
\begin{tabular}{l|ccc|ccc} 
\toprule
\multirow{2}{*}{Dataset} \tabletemplate \\
\midrule
CIFAR-10  & $0.778$ &$-0.778$ & $0.778$ & $-0.026$ & $0.118$ & $0.072$\\
CIFAR-100 & $0.133$ &$-0.118$ & $0.126$ & $0.162$ & $0.324$ & $0.227$\\
\bottomrule
\end{tabular}
}
\label{tab:corr_FTdataset}
\end{table}

\begin{figure*}[t]
\centering
\includegraphics[width=\textwidth]{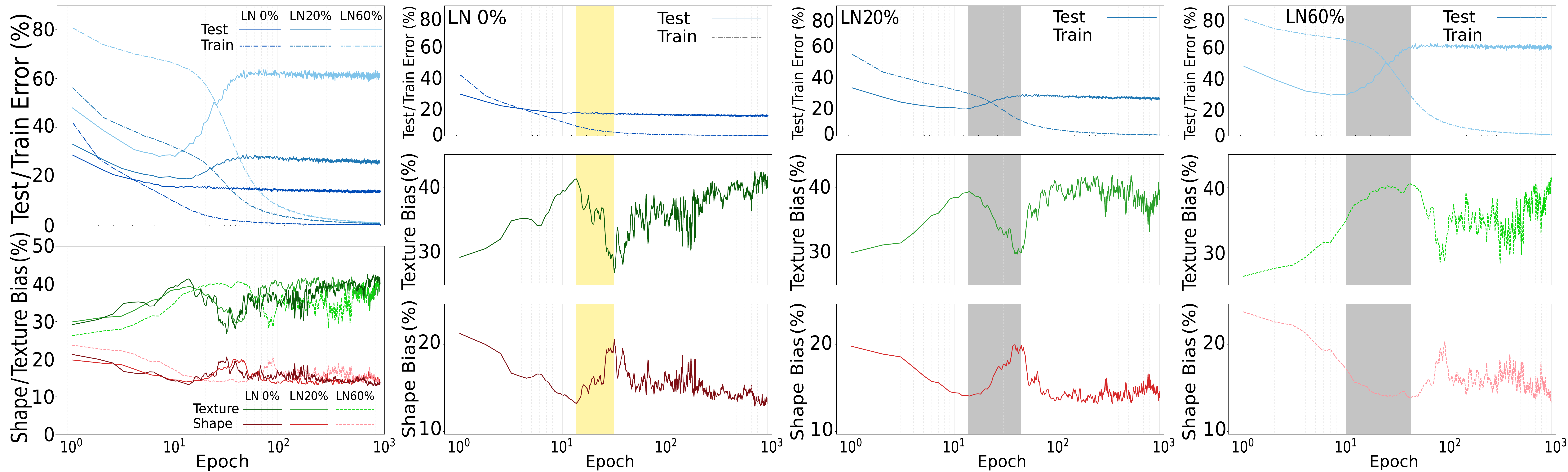}
\caption{Learning process under various label noise conditions. The label noise proportion $p$ is varied at 0, 0.2, and 0.6. While double descent is not observed in test error when $p=0$, we observed double descent/ascent phenomena in shape/texture bias.}
\label{fig:comp_ln}
\end{figure*}

\textbf{Parameter initialization.}
To investigate the effect of parameter initialization, this experiment compares training with randomly initialized weights and with weights pre-trained on ImageNet.
The results are shown in Fig. \ref{fig:comp_INSR}. We observed three phenomena.
First, in both cases, the texture/shape bias fluctuates synchronously with the double descent phenomenon of test error, as shown in the enlarged view on the right side of the figure.
Second, when random initialization is used, the absolute change of the bias is smaller. This is due to the residual effect of the Gaussian initialization.
Third, when training with pre-trained weights, the transition to Phase 2 is faster. This is because the training error decreases faster.
Overall, we observed similar phenomena in both cases.

\noindent{\textbf{Dataset.}}
This experiment examines the effect of changing the training dataset. We report results comparing the CIFAR-10 and CIFAR-100 datasets in Fig. \ref{fig:comp_IN_dataset} and the quantitative evaluation in Fig. \ref{tab:corr_FTdataset}.
We observed synchronization scores larger than 0.7 in Phase 1 and 2 on both datasets. The CIFAR-100 results show a corresponding correlation to the CIFAR-10 results: the correlation coefficients with respect to shape and texture bias were $0.778$ and $-0.778$ for CIFAR-10, respectively; correspondingly, they were $-0.689$ and $0.745$ for CIFAR-100.
This suggests that a similar dataset shows the same synchronization between test error and shape/texture bias.

\begin{table}[t]
\centering
\caption{
Correlation coefficients and synchronization scores for different label noise proportion $p$.
Results for $p=0$ are not reported because double descent was not detected.
}
\tablevspace
\scalebox{0.85}{
\begin{tabular}{c|ccc|ccc} 
\toprule
\multirow{2}{*}{$p$} \tabletemplate \\
\midrule
20\% & $0.778$ & $-0.778$ & $0.778$ & $-0.026$&$0.118$ & $0.072$\\
60\% & $-0.560$ & $0.598$ & $0.579$ & $0.122$ & $-0.142$ & $0.132$\\
\bottomrule
\end{tabular}
}
\label{tab:label_noise}
\end{table}

\noindent{\textbf{Label noise.}}
This experiment varies the proportion of label noise at 0\%, 20\%, and 60\%.
The results are shown in Fig. \ref{fig:comp_ln} and Fig. \ref{tab:label_noise}.
As the trend of the test error rate reveals, the magnitude of double descent increases as the label noise grows. However, when observing the shape/texture bias, there is a clear trend regardless of the label noise ratio. Especially in the texture bias, as the label noise increases, the shift from rising to declining seems to be delayed. This suggests that the progression of bias may slow down as label noise increases.
More interestingly, when the noise proportion is 0\%, double descent was not observed in test error, but we observed double descent/ascent phenomena of shape/texture bias (as colored in yellow in the figure). This indicates the possibility that there are  learning phases even in parts where double descent was thought not to occur.

\begin{figure*}[t]
\begin{minipage}[b]{0.32\linewidth}
\centering
\includegraphics[width=\linewidth]{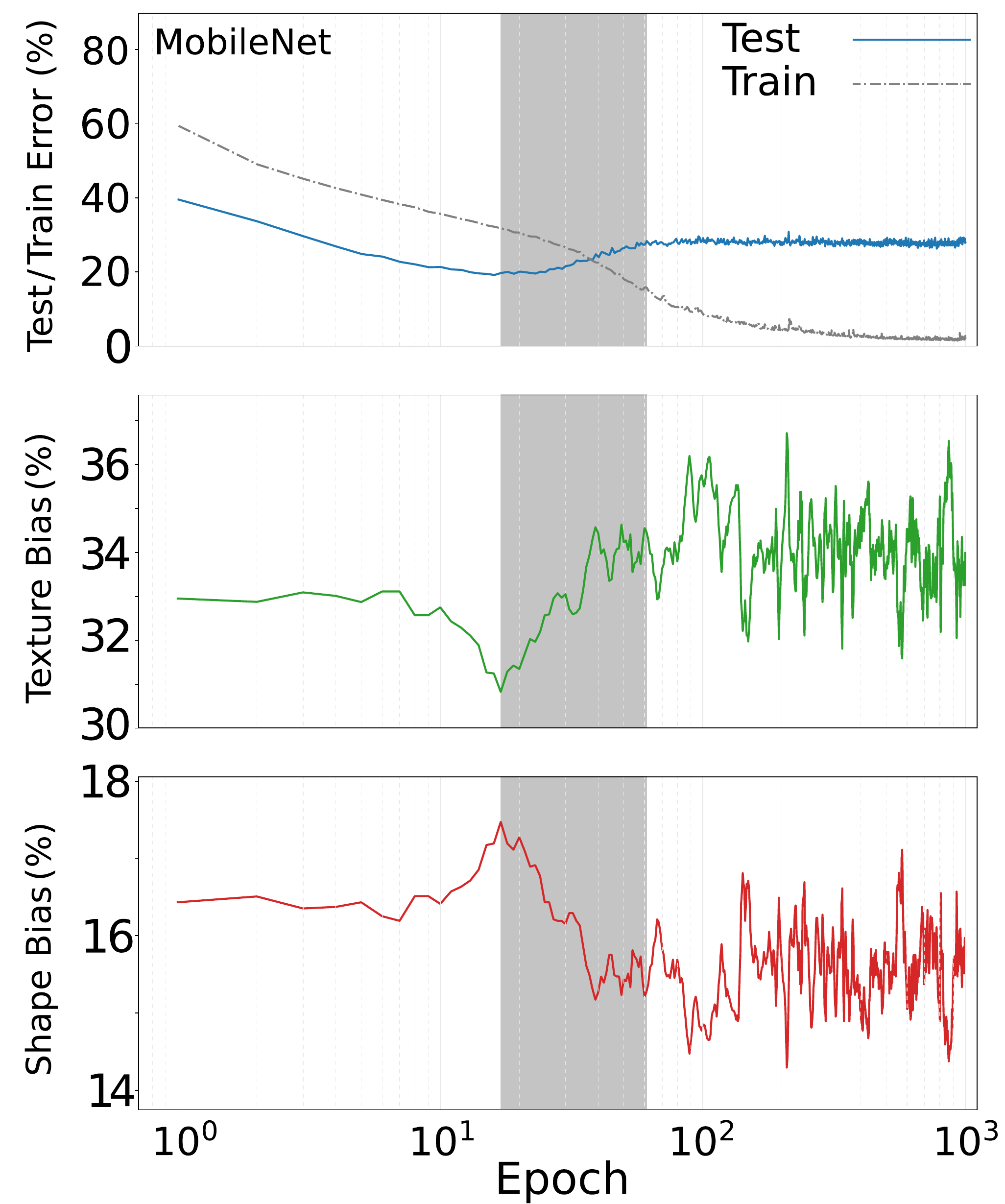}
\end{minipage}
\hfill
\begin{minipage}[b]{0.32\linewidth}
\centering
\includegraphics[width=\linewidth]{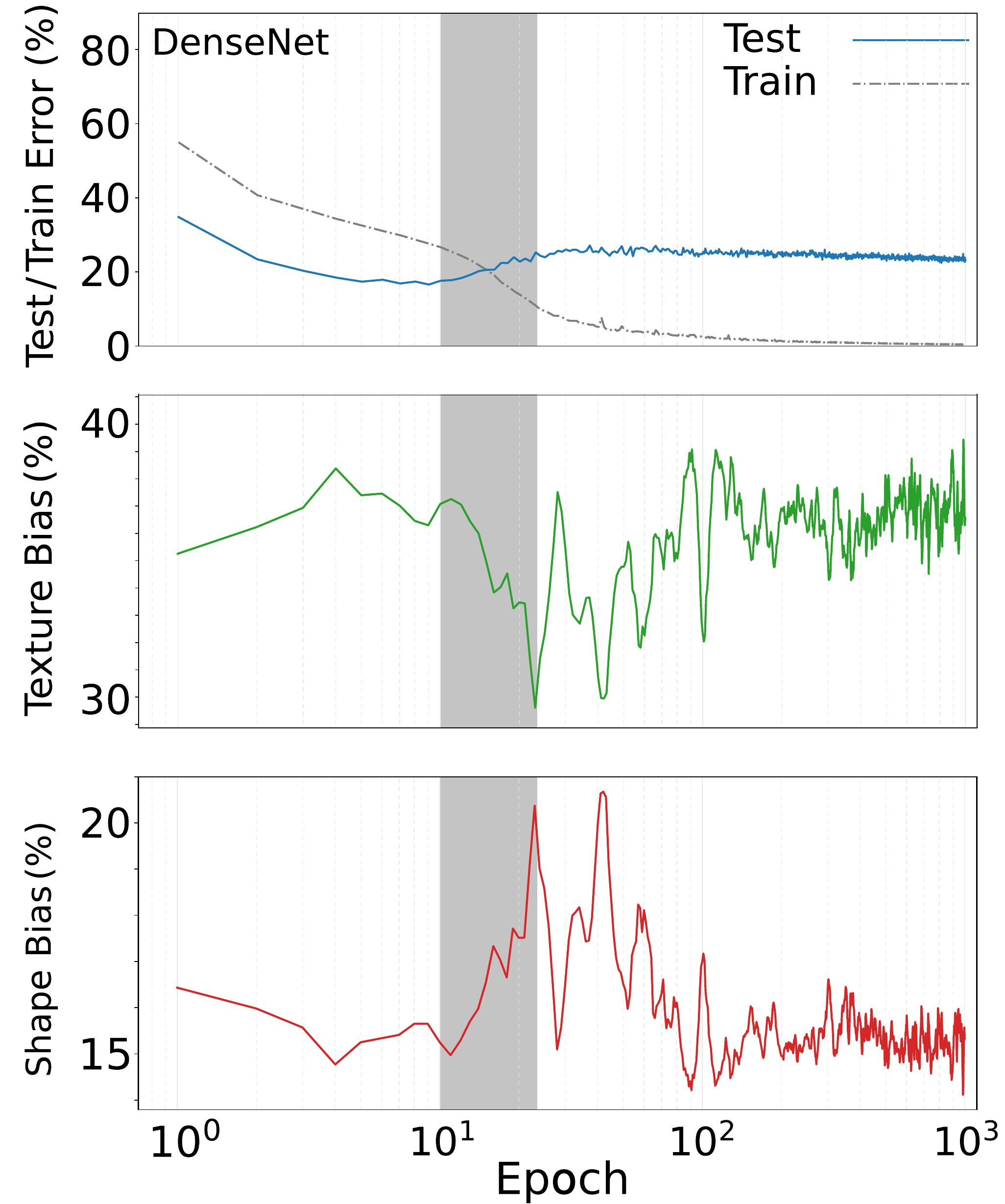}
\end{minipage}
\hfill
\begin{minipage}[b]{0.32\linewidth}
\centering
\includegraphics[width=\linewidth]{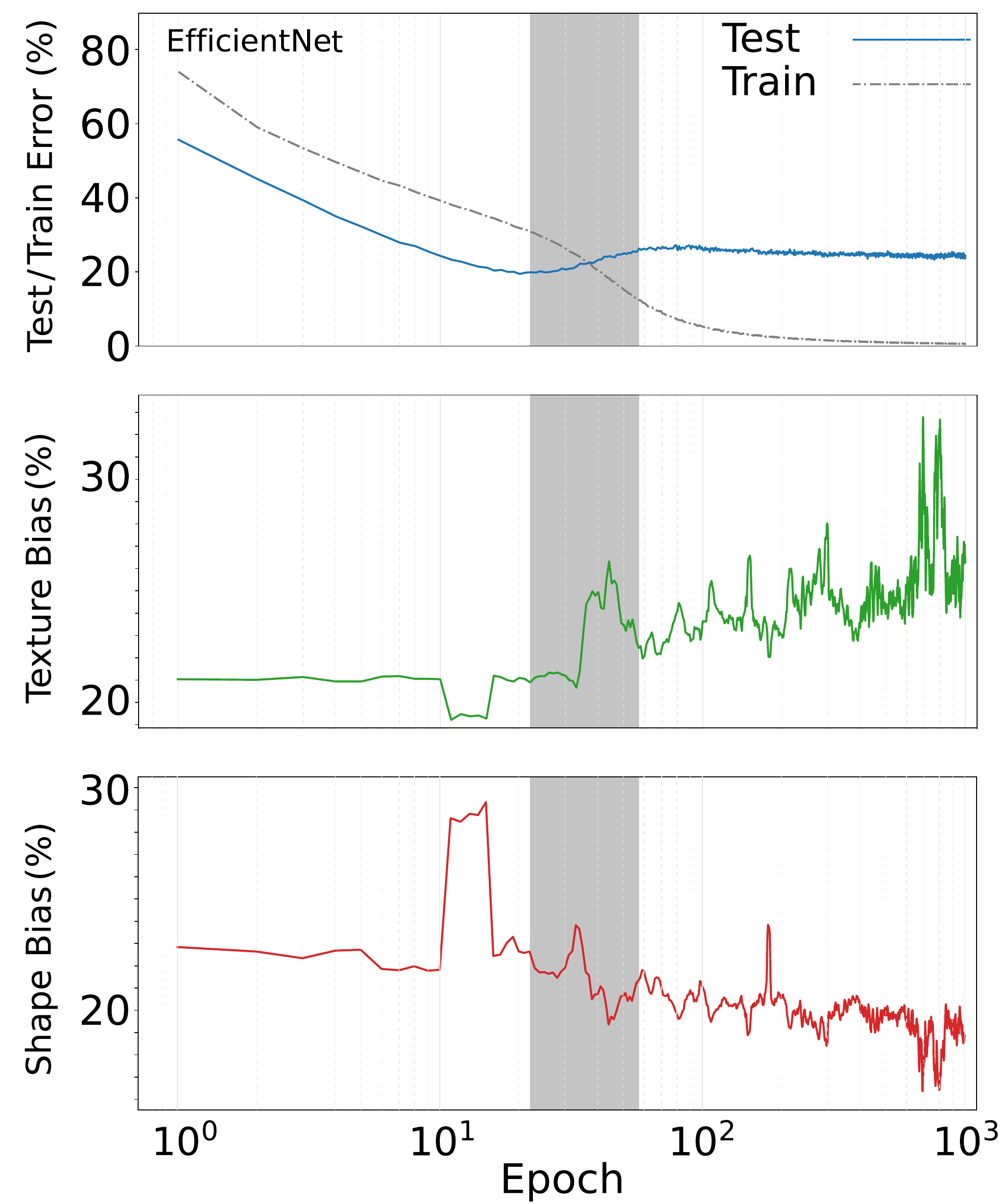}
\end{minipage}
\caption{Comparison of learning processes using different architectures. Left: MobileNetV2, Center: DenseNet121, Right: EfficientNetB0.}
\label{fig:architecture}
\end{figure*}

\noindent{\textbf{Architecture.}}
This experiment examines CNN architectures other than ResNet.
Specifically, we use MobileNetV2~\cite{mobileNetV2}, DenseNet121~\cite{DenseNet}, and EffecientNetB0~\cite{EffectiveNet}. We show the results in Fig.   \ref{fig:architecture} and Fig. \ref{tab:architecture}. The results reveal a moderate synchronization when using MobileNetV2 and DenseNet121. However, we observed no synchronization with EfficientNetB0.
This is because there is a spike at around epoch 10 in Phase 1 that significantly reduces the synchronization score.
This bias spike is currently an unknown phenomenon but could be related to the loss spike~\cite{zhang2023lossspike}, suggesting room for further discussion from the perspective of the stochastic learning process.

\begin{table}[t]
\centering
\setlength{\tabcolsep}{4.4pt}
\caption{Correlation coefficients and synchronization scores for different architectures.}
\tablevspace
\scalebox{0.85}{
\begin{tabular}{l|ccc|ccc} 
\toprule
\multirow{2}{*}{Architecture}  \tabletemplate \\
\midrule
MobileNet &\textminus0.506&0.511 & 0.509&\textminus0.016&0.036 & 0.026\\
DenseNet &0.326&\textminus0.322 & 0.324 & 0.293&\textminus0.289 & 0.291\\
EfficientNet &\textminus0.029&0.000 & 0.014&0.316     &\textminus0.343 & 0.330\\
\bottomrule
\end{tabular}
}
\label{tab:architecture}
\end{table}


\subsection{Layer-wise analyses and visualization}

Here, we conduct layer-wise analyses to investigate which layers are influenced by the bias. First, we analyze the shape/texture bias for hidden layers.
Second, we visualize the filters of the first convolution layer in each phase.

\begin{figure*}[t]
\centering
\includegraphics[width=\linewidth]{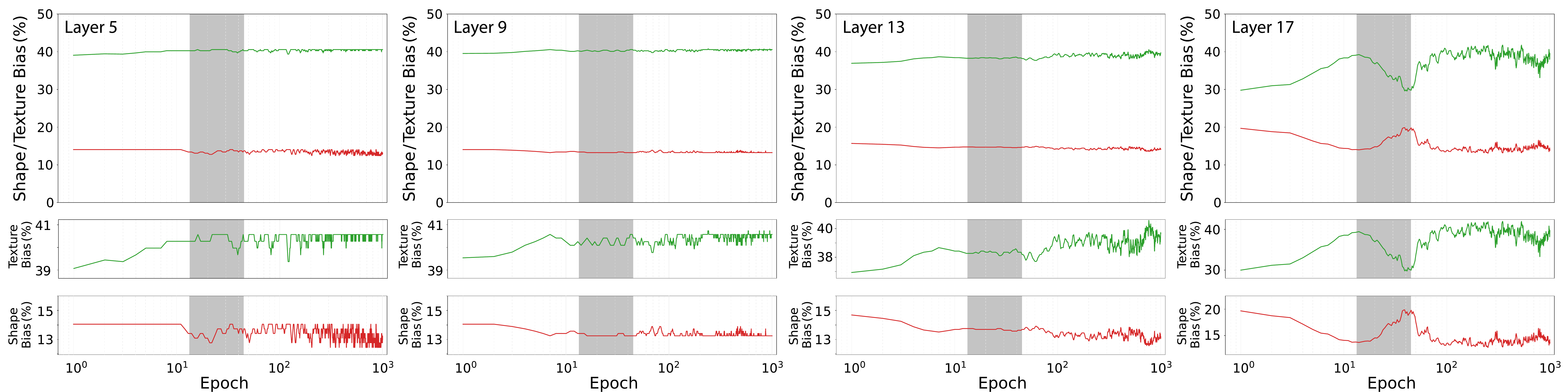}
\caption{
The shift of biases during the learning process in each layer consisting ResNet18. Using the convolutional layers of each block consisting ResNet18, including the final convolutional layer (17th layer), biases towards shape and texture are calculated using the same method as in \ref{sec;Quantifying the shape/texture biases of the model}. 
}
\label{fig:layer_ab}
\end{figure*}

\textbf{Shape/texture bias of hidden layers.}
Figure~\ref{fig:layer_ab} shows the shape/texture bias of 5th, 9th, 13th and 17th layers of ResNet18.
These results reveal that in convolutional layers, except for the 17th layer, there is no clear transition in shape/texture bias. This suggests that each layer may have unique inflection points and that the last few layers have a direct impact on test error.

\begin{figure*}[t]
    \centering
    \begin{subfigure}{0.23\textwidth}
        \centering
        \includegraphics[width=\linewidth]{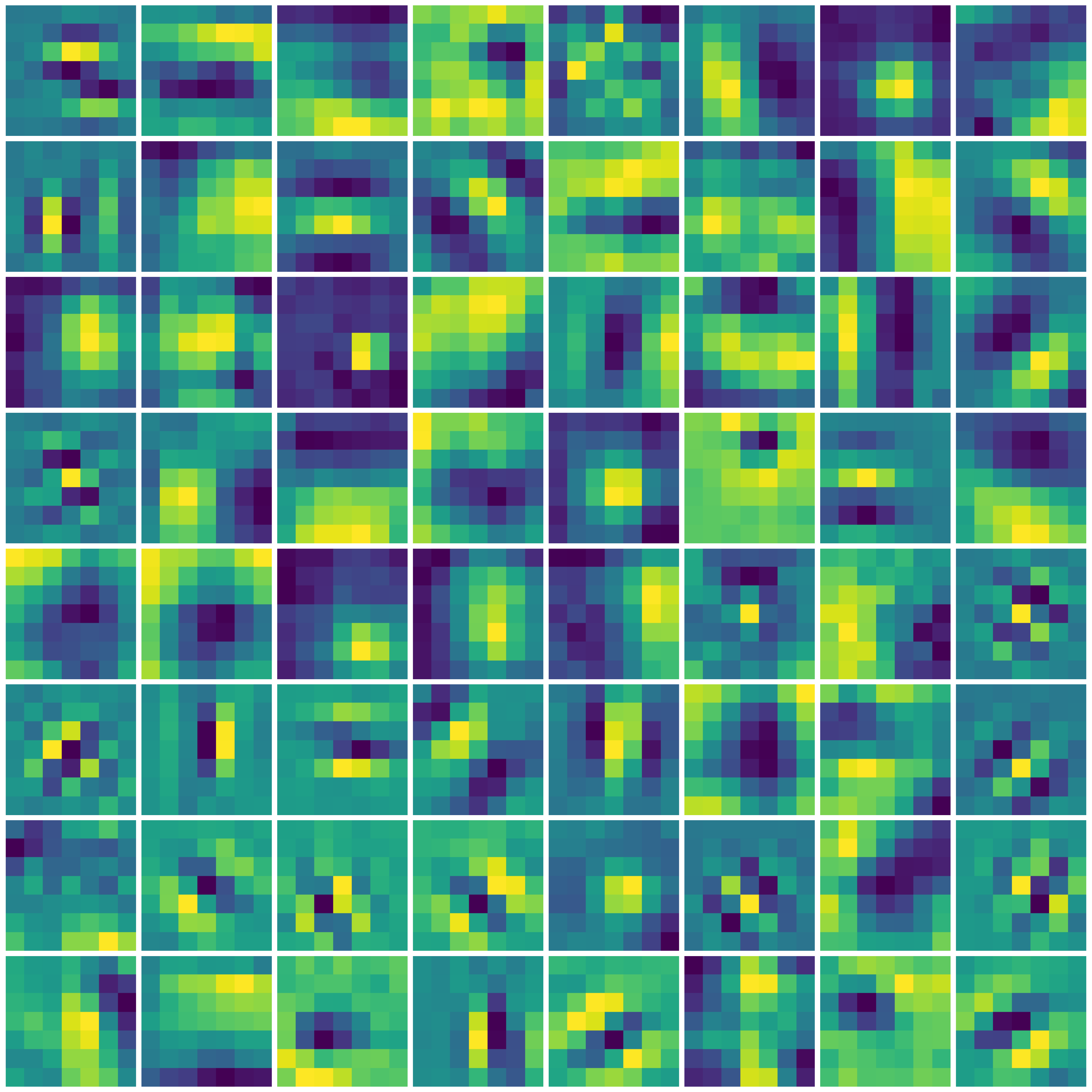}
        \caption{13th Epoch}
        \label{fig:13th}
    \end{subfigure}%
    \hfill
    \begin{subfigure}{0.23\textwidth}
        \centering
        \includegraphics[width=\linewidth]{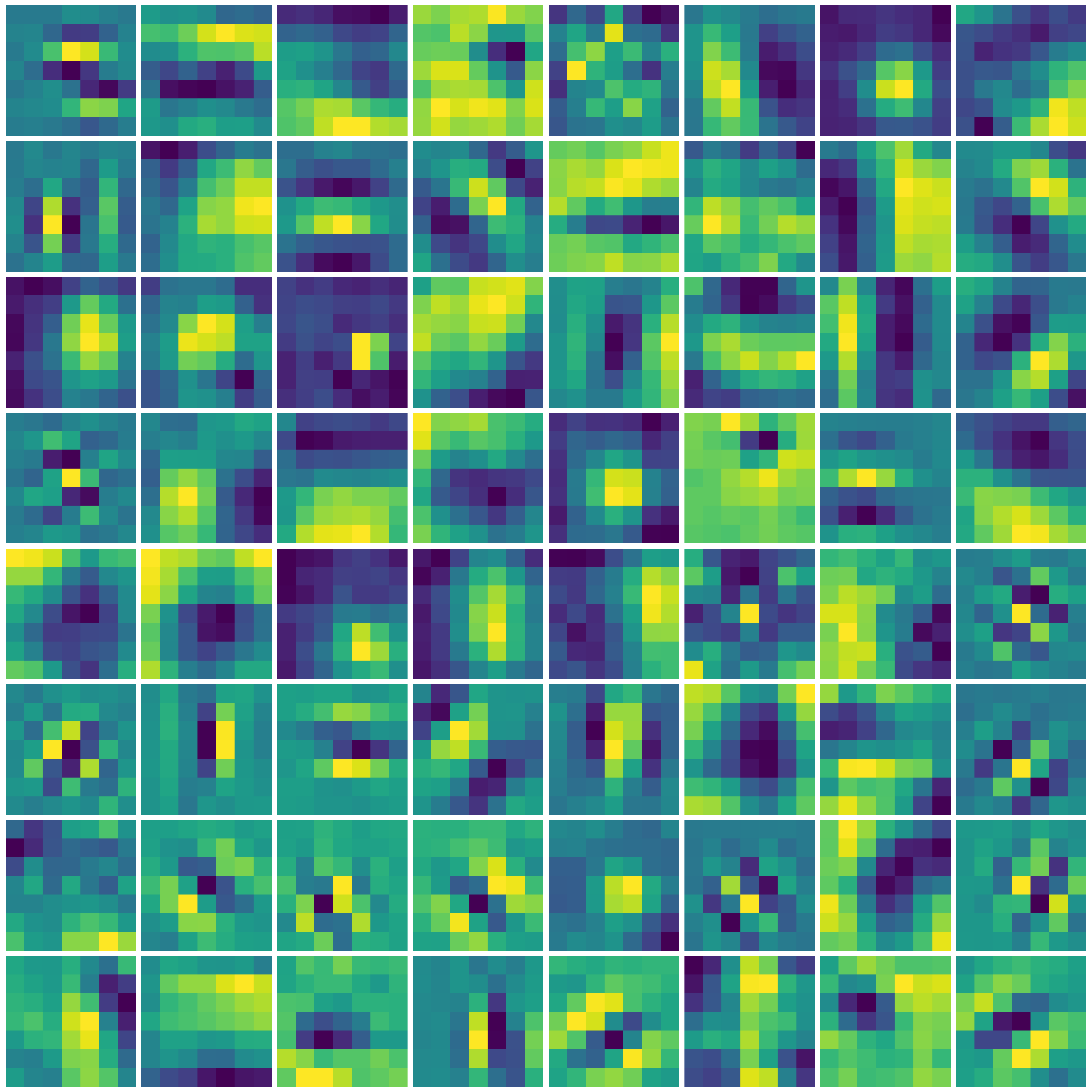}
        \caption{42nd Epoch}
        \label{fig:42nd}
    \end{subfigure}%
    \hfill
    \begin{subfigure}{0.23\textwidth}
        \centering
        \includegraphics[width=\linewidth]{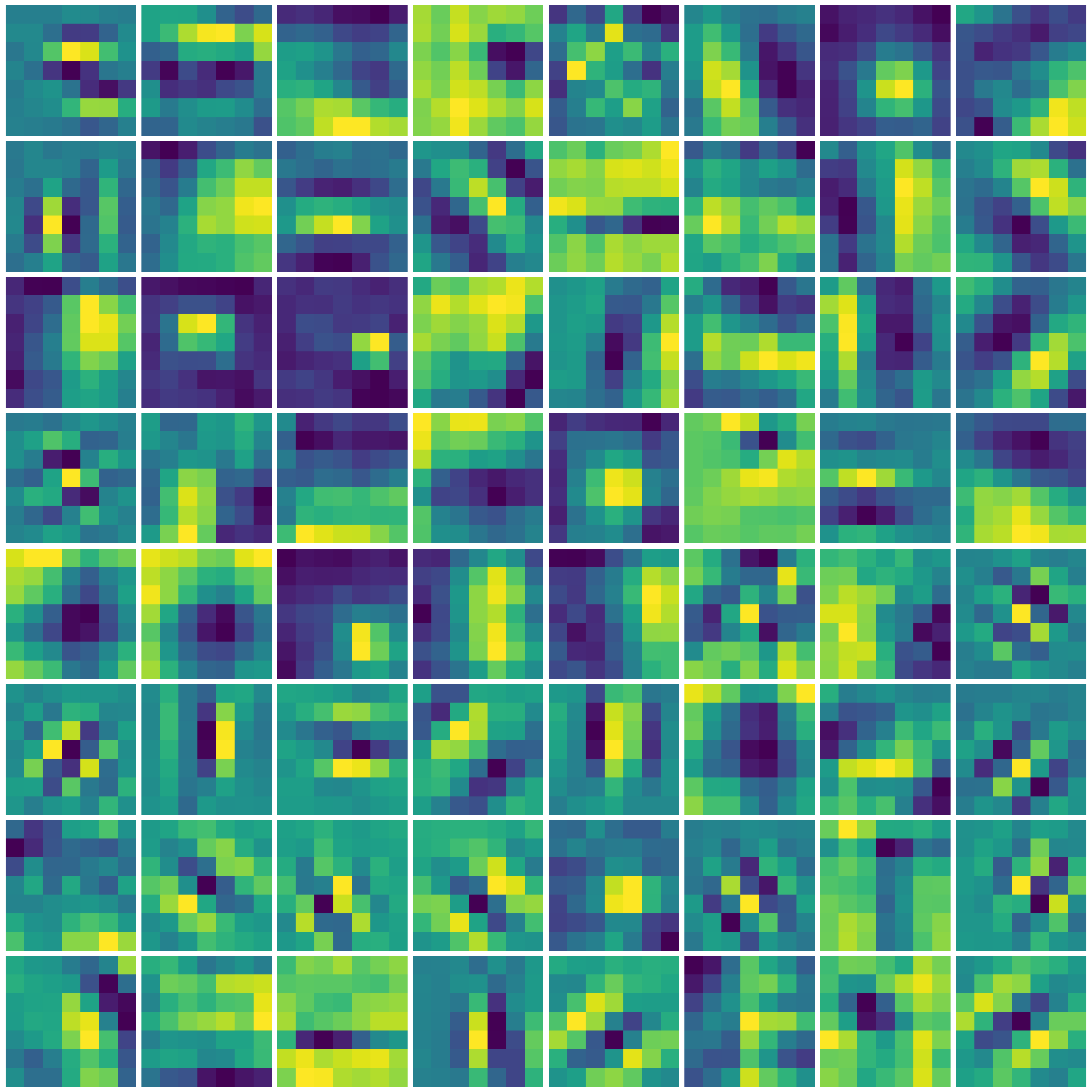}
        \caption{1,000th Epoch}
        \label{fig:1000th}
    \end{subfigure}
    \caption{
    Visualization of the 1st layer in the learning process: In the setting described in Fig. \ref{sec:Experimental setting}, we visualize the 1st layer in the Epoch (13th, 42nd Epoch) and the 1,000th Epoch, where the double descent is divided into 3 Phases. The 1st layer at the 1,000th Epoch is visualized.
    }
    \label{fig:visualized_kernel}
\end{figure*}

\textbf{Filter visualization.} 
Figure \ref{fig:visualized_kernel} visualizes the filters of the first convolution layer of ResNet18 at three points: the boundary between Phase 1 and Phase 2, the boundary between Phase 2 and Phase 3, and the 1,000th epoch. Although there are slight changes, the visualization confirms no significant changes in the filters. Considering the results of Fig. \ref{sec46}, this suggests that learning in shallow layers might not be affected by changes in shape/texture bias. This is consistent to the result of Fig. \ref{fig:layer_ab} where we observe a clear transition only on the 17th layer.


\section{Discussion}
In this study, we focused on shape and texture features in natural images and analyzed their relation to the double descent phenomenon. We found that under certain conditions, there is a double descent/ascent of shape/texture bias synchronized with a double descent of test error. In these conditions, there tends to be a correlation between the test error and shape/texture bias until the second descent in double descent of test error. However, this correlation disappears once the second descent begins. We discovered this trend by dividing double descent into three phases and quantitatively evaluating the correlations.
In subsequent experiments, we observed the synchronization of the bias and test error only in the final convolutional layer. This observation suggests that the deeper layers of the CNN may exhibit learning trends different from those of the intermediate layers.

Previous studies on double descent have proposed the hypothesis that it might be influenced by multiple features present in data. But do features like shape and texture directly cause the double descent phenomenon of test error? 
If such features were truly causing this phenomenon, then double descent behavior and bias towards shape and texture should show a more clear pattern. For example, shape bias might peak first, then decrease, followed by texture bias peaking. However, in reality, shape and texture biases show an inverse correlation.
Therefore, we believe that the phenomenon is more complex than it seems and that some learning tendencies causing double descent affect CNN's feature extraction tendencies, showing a correlation between double descent and bias progression.

From a practical perspective, when pre-trained on ImageNet, the possibility is suggested that test error may be minimized around epochs where bias is at its maximum or minimum. This implies that observing this bias could help determine the optimal number of training epochs. Furthermore, it was shown that factors causing double descent might also affect CNN's bias towards shape and texture features, especially in deeper layers.

In this study, we focused on the learning process of image features like shape and texture in CNNs and examined their relation to the double descent phenomenon under various conditions. There are many unexplored areas in deep learning, and double descent is one of them. This research highlights the importance of a deeper understanding of the deeper layers of deep learning, and it is considered to provide a promising direction for future research. However, given the complexity of neural network architecture, there are still unknown phenomena that need to be investigated in future research.

\section{Conclusion}
In this paper, inspired by previous studies on epoch-wise double descent, we focused on the relationship between image-specific features and double descent. We quantified the model's bias toward shape and texture to compare it with the test error. As a result, we discovered double descent/ascent of shape/texture bias synchronized with double descent of test error under Nakkiran's setting. Additionally, quantitative evaluations confirmed this correlation during the period from the initial decrease to the full increase in the test error. Further, we observed double descent/ascent of shape/texture bias in a condition without label noise, where double descent was thought not to occur. Layer-wise analyses deepened the understanding of the phenomena.

\textbf{Limitation and future work.}
To make a connection to previous studies, we chose Nakkarian's setting as a starting point for investigation. We believe this was the best choice for discussing the relationship between test error and shape/texture bias because it is the simplest setting where epoch-wise double descent of test error can be reproduced. However, this remains an investigation under large-scale training as future work. As recent computer vision studies focus more on large-scale training, it would be interesting to study the scaling of double descent.
Furthermore, extending bias quantification methods to modalities other than images, such as natural language and speech, is also challenging but promising to deepen the understanding of the double descent/ascent of bias in deep learning.
\section{Acknowledgement}


I would like to express my deepest gratitude to Mr. Yusuke Kyokawa, an alumnus of the Data Science and Machine Learning Laboratory at Tokyo Denki University, for his valuable discussions during the early stages of this research.

\bibliographystyle{splncs04}
\bibliography{ref}

\end{document}